\documentclass[letterpaper, 10 pt, conference]{ieeeconf}  

\IEEEoverridecommandlockouts                              

\usepackage{cite}
\usepackage{amsmath,amssymb,amsfonts}
\usepackage{algorithmic}
\usepackage{graphicx}
\usepackage{textcomp}
\usepackage{xcolor}
\usepackage{color, colortbl}
\usepackage{todonotes}
\usepackage{multirow}
\usepackage{siunitx}
\usepackage{subcaption}
\usepackage{float}
\usepackage{stfloats}
\usepackage{hyperref}
\usepackage{fancyhdr}
\usepackage{booktabs}

\hypersetup{
    colorlinks=true,
    linkcolor=black,
    filecolor=magenta,      
    urlcolor=cyan,
    pdftitle={Overleaf Example},
    pdfpagemode=FullScreen,
    }

\title{\LARGE \bf Collective PV-RCNN: A Novel Fusion Technique using Collective Detections for Enhanced Local LiDAR-Based Perception}

\author{Sven Teufel$^{1}$, Jörg Gamerdinger$^{1}$, Georg Volk$^{1}$ and Oliver Bringmann$^{1}$
\thanks{$^{1}$University of Tübingen, Faculty of Science, Department of Computer Science, Embedded Systems {\tt\small \{sven.teufel, joerg.gamerdinger, georg.volk, oliver.bringmann\} @uni-tuebingen.de}}%
}

\begin{document}

\maketitle
\thispagestyle{empty}
\pagestyle{empty}

\begin{abstract}
Comprehensive perception of the environment is crucial for the safe operation of autonomous vehicles. However, the perception capabilities of autonomous vehicles are limited due to occlusions, limited sensor ranges, or environmental influences. Collective Perception (CP) aims to mitigate these problems by enabling the exchange of information between vehicles. A major challenge in CP is the fusion of the exchanged information. Due to the enormous bandwidth requirement of early fusion approaches and the interchangeability issues of intermediate fusion approaches, only the late fusion of shared detections is practical. Current late fusion approaches neglect valuable information for  local detection, this is why we propose a novel fusion method to fuse the detections of cooperative vehicles within the local LiDAR-based detection pipeline. Therefore, we present Collective PV-RCNN (CPV-RCNN), which extends the PV-RCNN++ framework to fuse collective detections.
Code is available at \url{https://github.com/ekut-es}
\end{abstract}

\section{Introduction}
For the safe operation of intelligent vehicles (IVs) in complex driving scenarios, a comprehensive perception of the environment is crucial. 
However, in real-world driving, an IV can only partially perceive the environment due to limited sensor ranges and occlusion by other road users or infrastructure \cite{volk_environment-aware_2019}. Especially in harsh environmental conditions such as rain, snow, or fog, the sensing capabilities of an IV can be severely degraded \cite{teufel2022simulating, teufel2023enhancing}. Collective Perception can mitigate these problems by sharing information about perceived objects between vehicles. A major challenge in CP is the fusion of the received data into the local environment model. Several approaches for CP exist, which differ mainly in the type of data exchanged between vehicles. These approaches can be divided into three classes: early, late and intermediate fusion. 

In early fusion, raw sensor data are shared by the IVs. This allows the IVs to perform detections on the basis of all available environmental information. Therefore, to be fused, the sensor data from other IVs must be aligned with the local sensor data. However, only a spatial alignment is feasible, since there is no information about dynamic objects, a temporal alignment is difficult. This approach is considered the most accurate since there is no loss of information, but it is not practical due to bandwidth limitations in communication~\cite{yuan2022keypoints}.

 For late fusion, preprocessed information about detected objects is used. The IVs exchange states (position, velocity, acceleration) of their detected objects. The received detections are then fused with the local detections. This approach requires only a low bandwidth \cite{caillot2022survey} and there already exist communication standards like the Collective Perception Message (CPM) to exchange object states~\cite{CPM}. However, due to information loss during preprocessing this approach is considered less accurate than the early fusion~\cite{caillot2022survey}.
Intermediate fusion combines the advantages of both, early and late fusion by exchanging features from the local object detector between the vehicles.
In the case of LiDAR-based detection, these could be voxel features \cite{chen2019f} or point features \cite{yuan2022keypoints}. This method uses less bandwidth than early fusion while being more accurate than late fusion, but all IVs need exactly identical detection pipelines to use the shared features, making this approach impractical.

In this work, we propose a novel fusion method called Collective PV-RCNN that relies solely on the detections of other IVs and directly fuses them within the local LiDAR-based object detection pipeline. We have extended the implementation of the state-of-the-art LiDAR-based object detector PV-RCNN++ \cite{shi2023pv} from OpenPCDet \cite{openpcdet2020} to include collective detections.
This approach uses valuable information from cooperative vehicles for the local detection, which is neglected in current late fusion approaches. 
This approach provides a certain level of validation of the received detections in the local sensor's field of view, as they are not directly fused into the local environment model.

In this work we give an overview on related research in Section II and present several methods to incorporate collective detections into the local detection pipeline in Section III. Afterwards we present the conducted experiments together with the dataset, that was used for the evaluation, in Section IV. We provide our results in Section V and conclude our work in Section VI, which also gives an outlook on future research.

\section{Related Work}
\label{rel_work}

\subsection{Object Detection}
Shi et al. \cite{shi2020pv} proposed the Point-Voxel R-CNN (PV-RCNN), which is a two-stage LiDAR-based object detector that uses both a voxel-based and a point-based representation. In the first stage, the point cloud is voxelized and then processed by a 3D sparse Convolutional Neural Network (CNN) as presented in SECOND \cite{yan2018second}, followed by a 2D Region Proposal Network (RPN) that generates 3D bounding box proposals. The second stage then samples a fixed number of points from the point cloud, called keypoints, using Farthest Point Sampling (FPS). An adopted set abstraction as in PointNet++ \cite{qi2017pointnet++}, called voxel set abstraction, is then used to aggregate the sparse voxel features from the 3D CNN to the keypoints along with 2D bird's eye view features from the RPN. These keypoint features are then aggregated in the RoI-grid pooling module by a PointNet-based set abstraction \cite{qi2017pointnet} into grid points that are sampled within the bounding box proposals. These fixed-size RoI grid features are then used by a Multi Layer Perceptron (MLP) to refine the 3D bounding box proposals and predict a confidence score. An improved version of PV-RCNN, called PV-RCNN++, has also been proposed by Shi et al. \cite{shi2023pv}. In PV-RCNN++ there are two major improvements over PV-RCNN. First, Farthest Point Sampling is replaced with Sectorized Proposal Centric (SPC) sampling. SPC sampling uses FPS to sample keypoints only within a radius around bounding box proposals, resulting in more representative and less background points sampled compared to using FPS on the entire point cloud. The second improvement was to replace the set abstraction in the VSA and RoI grid pooling with the VectorPool Aggregation, which is computationally cheaper than the set abstraction. The VectorPool aggregation generates a feature vector for each keypoint by aggregating local voxel features. PV-RCNN++ could slightly improve the results of PV-RCNN on common benchmark datasets while significantly reducing the resource consumption, making it about three times faster than PV-RCNN \cite{shi2023pv}.

\subsection{Early Fusion}
For the early fusion approach, raw sensor data is exchanged between vehicles.
Chen et al. \cite{chen2019cooper} evaluated the performance of early fusion on their own dataset as well as on the KITTI \cite{geiger2012we} dataset by aligning LiDAR point clouds from different time steps and performing an object detection. Their findings show a significant improvement in detection performance for the evaluated scenarios. However, the datasets used for evaluation are not dedicated to evaluate CP approaches since they contain only ego vehicle recordings. They state that a transmission of raw point clouds with existing vehicular networks is possible. However, they preprocess and transmit only parts of the point cloud of a low resolution LiDAR sensor. Moreover, an insufficient sensor data rate of \SI{1}{\hertz} was used.
An investigation on fusion strategies for collective perception with infrastructural LiDAR sensors was conducted by Arnold et al.~\cite{arnold2020cooperative}. They demonstrated that the early fusion outperforms the late fusion in a T-junction and a roundabout scenario. In both cases, the 3D average precision could be increased by up to 17 percentage points (p.p.). However, stationary sensors can use wired communication; hence, it can not be compared to to V2X communication.

\subsection{Late Fusion}

Late fusion uses preprocessed data from other vehicles to enhance vehicle-local perception. A simple way to realize late fusion is to use detected bounding boxes from multiple detectors and weight them according to the detection confidence. Solovyev et. al~\cite{solovyev2021weighted} introduced a Weighted Mean Fusion (WMF) for local image-based detection.
Houenou et. al~\cite{houenou_track--track_2012} and Müller et. al~\cite{muller_generic_2011} applied WMF for multisensor fusion of vehicle-local sensor data. Aeberhard~\cite{aeberhard_object-level_2017} extended this approach to a WMF for collective perception.
The advantage of WMF is that it requires very little data, so that the communication channel of collective perception is not too busy. However, WMF, does not consider correlation between data and is less precise compared to early fusion.

Another way of late fusion is to use an adapted Kalman filter. The collectively perceived tracks are considered as measurements and integrated into the local environmental model. Approaches using Kalman filter based collective perception are presented in~\cite{allig2019,volk_environment-aware_2019,gabb2019infrastructure,volk2021}. These approaches additionally use tracking information in contrast to simple weighted mean fusion, which is based on detection information. The fused data still contains less information compared to early fusion. Therefore, the result is less precise.

Late fusion approaches based on covariance intersection~(CI) are presented in ~\cite{aeberhard_object-level_2017,gabb2019infrastructure,ajgl_covariance_2018} and rely on the cross-covariance for fusing the collectively perceived track information. Volk et al. \cite{volk2022environment} further improved CI fusion by additional validation of the perceived information.
In contrast to CI, Information Matrix Fusion~(IMF) as used in~\cite{aeberhard_object-level_2017,gabb2019infrastructure} only considers new information of perceived tracks to avoid correlation.
Both CI and IMF respect the data correlation to gain improved fusion results and incorporate more information compared to the weighted mean fusion. 
Similar to the previous late fusion approaches, there is an information loss by using preprocessed data in comparison to having access to the raw data as in early fusion approaches. Hence, it is not possible to achieve the same perception quality as not all information is present.

\subsection{Intermediate Fusion}
The intermediate fusion approach is based on the exchange of features from the local object detection.
Chen et al. \cite{chen2019f} extended their early fusion approach by sharing voxel or spatial features from the VoxelNet \cite{zhou2018voxelnet} to reduce the amount of transmitted data. The approach of sharing voxel features performed similar to the raw data sharing while reducing the transmitted data slightly. With the sharing of spatial features, the transmitted data was significantly reduced, but the detection performance also dropped significantly. Similar to the sharing of voxel features, Bai et al. \cite{bai2021pillargrid} shared pillar features from the PointPillars \cite{lang2019pointpillars} detector. They used a simulated dataset generated with CARLA \cite{CARLA}, that included a vehicle and a roadside unit equipped with one LiDAR sensor each. Similar to Chen et al.~\cite{chen2019f}, their results show a significant improvement in detection performance compared to local-only detection. The sharing of keypoint features of the PV-RCNN detector was evaluated by Yuan et al. \cite{yuan2022keypoints} on their own simulated dataset. Their results show that sharing keypoint features can significantly reduce the amount of shared data compared to sharing entire feature maps, while achieving better detection performance. Another attentive intermediate fusion was proposed by Xu et al.~\cite{xu2022opv2v}. They use metadata sharing to build local groups of vehicles, transforming their point clouds to the ego vehicles space and then extract features that are encoded and shared with the ego vehicle.
Moreover, they conducted an evaluation of early, intermediate and late fusion using different object detectors. For all detectors, the early fusion achieved a higher AP than the late fusion; however, their proposed intermediate fusion outperformed the early fusion by up to 10 p.p.~\cite{xu2022opv2v}.
All these approaches could show a significant improvement in detection performance while reducing the amount of data transmitted, but none of them address the aforementioned interchangeability problem. In all approaches, the identical detector is used for all vehicles, which makes these approaches impractical.

\section{Collective Detection Fusion}

\label{sec:detection_fusion}
\begin{figure*}
    \centering
    \includegraphics[width=\textwidth]{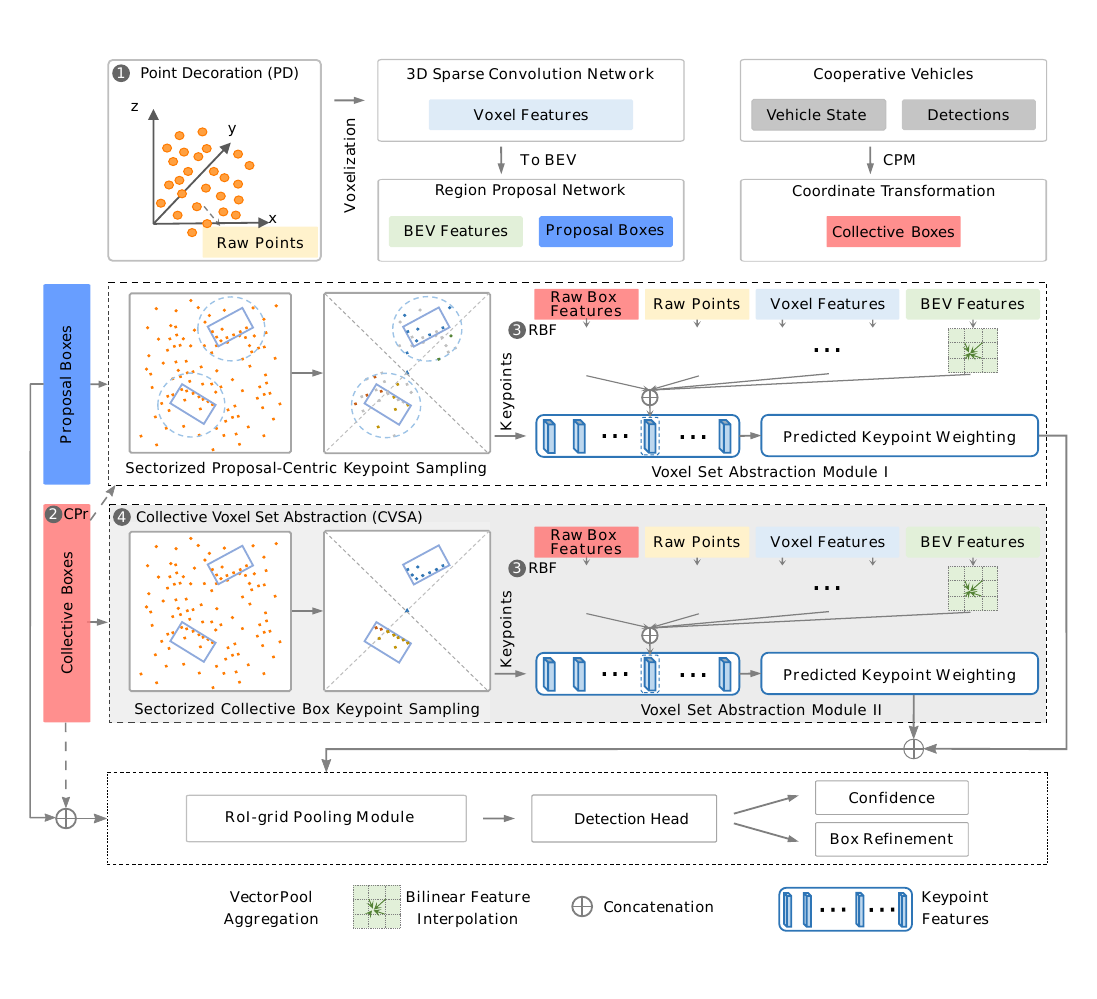}
    \caption{The overall architecture of the extended PV-RCNN++ \cite{shi2023pv} detector. We propose four independent extensions to the PV-RCNN++ detector, which can be used individually or in combination. \raisebox{.5pt}{\textcircled{\raisebox{-.9pt} {1}}} Point Decoration (PD): The input points are decorated with additional point features, determined by the collective detections, in which the points are located. \raisebox{.5pt}{\textcircled{\raisebox{-.9pt} {2}}} Collective Proposals (CPr): The collective detections are used as additional region proposals, either as additional input to the sectorized proposal centric keypoint sampling module (CPr SPC) or the RoI-grid pooling module (CPr RoI-grid). \raisebox{.5pt}{\textcircled{\raisebox{-.9pt} {3}}} Raw Box Features (RBF): The raw bounding boxes of the collective detections are used as additional input to the voxel set abstraction module. \raisebox{.5pt}{\textcircled{\raisebox{-.9pt} {4}}} Collective Voxel Set Abstraction (CVSA): A second keypoint sampling module is added, which samples keypoints within the collective detection, followed by a second voxel set abstraction module. Figure adapted and extended from \cite{shi2023pv} licensed under a Creative Commons Attribution 4.0 International License.}
    \label{fig:pvrcnn}
\end{figure*}

We propose a novel approach to improve local LiDAR-based object detection of IVs by incorporating cooperative vehicle detections into the local object detection. Our approach extends the PV-RCNN++ \cite{shi2023pv} detector architecture. We rely solely on detections exchanged via CPMs between cooperative vehicles. Before fusion, these detections need to be transformed to the local coordinate system. Each detection $D$ consists of the 3D object position $(x,y,z)$, the object dimensions $(w,h,l)$, the object orientation $\Theta$, the object class, and a confidence score.
In this section, we present four different methods to incorporate the received detection of other IVs into the local detection pipeline. Each of these methods is an independent extension to the detector and can be used individually as well as in combination with the other methods. An overview of the complete architecture of our approach is given in Figure~\ref{fig:pvrcnn}.

\subsection{Point Decoration (PD)}
One way to incorporate collectively detected objects into local perception is to decorate the points of the locally perceived point cloud. Each point $P(x,y,z,i)$, where $x,y,z$ is the 3D position and $i$ is the intensity, is decorated with an additional point feature $\Sigma_{conf}$, which is the sum of the confidences of all collective detections containing $P$. Thus, each $P$ becomes $P'(x,y,z,i,\Sigma_{conf})$. We only add one additional point feature since we are only focusing on one class in the evaluation. For multi-class detection, this approach can be extended analogously by adding one point feature per class to the points. Since only the input data is modified, this approach is applicable to all LiDAR-based object detector architectures.

\subsection{Collective Proposals (CPr)}
Another way to incorporate collective detections to improve the local perception is to use them as region proposals for the detection head, which predicts the confidences and refines the region proposals for the final detection. 
The region proposals require an objectness score, which is usually predicted by the region proposal network. For this, the confidence of the collective detection can be used. Since every two-stage detector uses region proposals, this approach can be applied to all two-stage detectors. In the case of PV-RCNN++, the bounding box proposals are used as input to the RoI-grid pooling module where they are used to aggregate local features into a single local vector representation. This local vector representation is then used for confidence prediction and proposal refinement by MLP layers. This variant will be referred to as CPr RoI-grid. In PV-RCNN++, the bounding box proposals are also used to sample keypoints that are within a certain radius of the proposals. These keypoints are the link between the point-voxel representation and are used to aggregate multi-scale scene features. Therefore, the keypoint sampling can strongly influence the overall performance of the detection. In this case the collective detections can also be used as input for the SPC keypoint sampling module to sample keypoints from the local point cloud at the locations of the collective detections. We will refer to this variant as CPr SPC.

\subsection{Raw Box Features (RBF)}
\label{RBF}
The Voxel Set Abstraction (VSA) module uses the aforementioned keypoints to aggregate features from different sources, in the original PV-RCNN++ architecture. The features are raw point features from the input point cloud, sparse voxel features from the 3D sparse convolution network, and bird's eye view (BEV) features from a 2D BEV feature map. The point and voxel features are aggregated using the vector pool aggregation, where local features are generated by dividing the 3D space around a keypoint into voxels and encoding them with separate kernel weights. For the BEV features, bilinear interpolation is used instead of vector pool aggregation. In addition to these feature inputs, we propose to also use the raw box features of the collective detections $D$ as feature inputs in the VSA module in the same way as the raw point features. The raw box features are then concatenated with the other feature inputs after the vector pool aggregation.

\subsection{Collective Voxel Set Abstraction Module (CVSA)}
Since the keypoint sampling strongly influences the detection performance, we propose a different keypoint sampling method called Sectorized Collective Box (SCB) keypoint sampling, where we sample keypoints only inside the bounding boxes of collective detections. This is in contrast to the SPC keypoint sampling, where the keypoints are sampled within a radius around the region proposals. This allows us to sample points that belong to objects more densely, since the collective detections are expected to be more accurate than the region proposals. In the same way as in the SPC keypoint sampling, the keypoints are sampled from sectors in order to accelerate the sampling process. Therefore, the region around a proposal is divided into scene centered sectors and the points are sampled from these sectors in parallel. To aggregate the keypoint features for the keypoints sampled with the SCB sampling, we added a second VSA module as shown in Fig. \ref{fig:pvrcnn}. The raw box features as described in section \ref{RBF} can also be used as feature input in this VSA module. The keypoint features of both VSA modules are then concatenated as input for the RoI-grid Pooling Module.

\newpage

\begin{table}
\centering
\caption{3D average precision results for the separate proposed fusion methods}
\label{tab:results_single}
\renewcommand{\arraystretch}{1.5}
\begin{tabular}{@{}lll@{}}
\toprule
Method                                     & AP@IoU$_{0.7}$ & AP@IoU$_{0.5}$  \\ \midrule \midrule
Baseline                                   & 34.26 & 74.15 \\ \midrule
Raw Box Features (RBF)                     & 39.45 & 76.61 \\
Point Decoration (PD)                      & 39.88 & 75.81 \\
Collective Voxel Set Abstraction (CVSA)    & 45.47 & 75.62 \\
Collective Proposals (CPr) RoI-grid        & 69.25 & 83.80 \\
Collective Proposals (CPr) SPC             & \textbf{69.59} & \textbf{84.08} \\
\bottomrule
\end{tabular}
\end{table}

\section{Evaluation}
\label{sec:eval}
\subsection{Dataset}
To train the PV-RCNN++ detector and evaluate the proposed approach we generated a synthetic collective perception dataset. For this, we used the CARLA simulator~\cite{CARLA} and the RESIST framework~\cite{resist}. CARLA allows to simulate multiple sensors on cooperative vehicles and provides comprehensive ground truth (GT) information. The RESIST framework is connected to CARLA via the C\texttt{++} API and allows to store sensor data and GT in KITTI format, which enables to use existing detection frameworks.

For training and testing a highway section with four lanes on the CARLA map Town04 is used. The ego vehicle and three cooperative vehicles are equipped with one \SI{360}{\degree} LiDAR sensor with 64 channels (\SI{10}{\hertz}, 1.3 $M$ points per second, \SI{120}{\metre} sensing range, \SIrange{-24.9}{2.0}{\degree} vertical FOV, $\pm$ \SI{0.02}{\metre} error). CARLA was used in synchronous mode, such that the sensor frequencies are synchronized.  Further, up to 25 randomly spawned vehicles per scenario are used as traffic to be detected. For training of the detector we used the recordings from the cooperative vehicles and for testing we used the ego vehicle recordings.

\subsection{Experiments}
In order to evaluate our proposed fusion approach, we used the standard PV-RCNN++ architecture to detect objects in the cooperative vehicle recordings. These detections are then shared with the ego vehicle. Since we want to focus on the evaluation of the fusion of collective detections, we assume no communication delay, so that all detections of the cooperative vehicles are available in each time step (i.e. for each of the synchronized sensor frames). We trained the CPV-RCNN using the HANNAH framework \cite{gerum2022hardware} for each of the proposed fusion approaches, except for the collective proposals, since they are only used at test time. We report the average precision for 3D detection with the 40 recall points interpolation, as used in the KITTI benchmark\cite{geiger2012we}, with Intersection over Union of $0.7$ (IoU$_{0.7}$) and $0.5$ (IoU$_{0.5}$) as true positive threshold. Furthermore, we compare our approach with a common late fusion approach, therefore we used the Hungarian Matching (HM) \cite{kuhn1955hungarian} together with a Weighted Box Fusion (WBF) \cite{solovyev2021weighted}. The detections of the cooperative vehicles are matched based on the euclidean distance to the detections of the ego with a maximum matching distance of \SI{2}{\metre}. The unmatched detections are then matched with the other cooperative detections in the same manner. For fusion, the detections are weighted by their confidence score and then fused using the mean of all box features. Since we focus on fusion of the detections, we do not apply a tracking algorithm.
Each of the proposed methods is evaluated separately as well as in combination with the other methods. Additionally, we evaluate the combinations of our fusion methods together with the aforementioned late fusion approach. 

\section{Results}
\label{results}

The evaluation is conducted as presented in Section ~\ref{sec:eval} using the mentioned dataset. An overview of the results for our proposed fusion methods is shown in Table~\ref{tab:results_single}.

\begin{table}
\centering
\caption{3D average precision results for different combinations of our methods}
\label{tab:results_combinations}
\renewcommand{\arraystretch}{1.5}
\begin{tabular}{@{}lll@{}}
\toprule
Method                                   & AP@IoU$_{0.7}$ & AP@IoU$_{0.5}$  \\ \midrule \midrule
Baseline                                 & 34.26          & 74.15           \\ \midrule
RBF + CPr RoI-grid                       & 72.58          & 82.88           \\
RBF + CPr SPC                            & 72.92          & 83.35           \\
PD + CPr RoI-grid                        & 73.90          & 91.25           \\
PD + CPr SPC                             & 74.02          & \textbf{91.42}  \\
CVSA  + CPr RoI-grid                     & 78.49          & 89.20           \\
CVSA  + CPr SPC                          & \textbf{79.02} & 89.33           \\
PD + CVSA + CPr RoI-grid                 & 72.36          & 87.27           \\
PD + CVSA + CPr SPC                      & 67.66          & 81.90           \\
\bottomrule
\end{tabular}
\end{table}

As baseline, a local perception with the standard PV-RCNN++ architecture is used. The baseline achieved an AP of \SI{34.26}{\percent} for AP@IoU$_{0.7}$ and \SI{74.15}{\percent} for AP@IoU$_{0.5}$. First, we evaluated the methods proposed in Section~\ref{sec:detection_fusion} individually. RBF and PD achieved about \SI{39}{\percent} AP, which is a minor increase compared to the baseline. This could be caused by poor region proposals which leads to an improper keypoint sampling in the regions of collective detections, since both methods do not necessarily result in region proposal of the collective detections. Thus, the increase of the AP is limited for these methods. With an increase in AP of about 11~p.p., the CVSA method showed a significant improvement. Similar observations and AP increases can be observed for AP@IoU$_{0.5}$. 
However, the CVSA method was outperformed by both approaches of Collective Proposals (SPC and RoI-grid) with an AP of about \SI{69}{\percent}, which is an increase of 35~p.p. or about \SI{100}{\percent} compared to the baseline. As for an IoU of 0.7, the CPr strategy achieves a significant improvement compared to the aforementioned strategies using an IoU threshold of 0.5. However, the percentage increase is only about \SI{15}{\percent} which can be traced back to the higher baseline AP. The CPr methods directly affect the region proposals of the two stage detector and, in the case of the SPC variant, also improve the keypoint sampling. Based on the improved region proposals, the detection head is able to predict the confidences and adjust the region proposals more precisely for the final detection, which strongly affects the detection quality.

In order to further improve the local detection, we investigated the perception capabilities using different combinations of the proposed methods. The results of the combined methods are given in Table \ref{tab:results_combinations}. All combined methods achieved higher AP scores compared to the individual application of the methods. Combining both of the CPr methods with RBF or PD lead to an AP of about \SI{73}{\percent} which is an improvement of 4 p.p. compared to the best individual method (CPr). The minor improvement of RBF and PD compared to the baseline can be transferred to the combined method as the methods concern different parts of the architecture. The best result for AP@IoU$_{0.5}$ was achieved by the PD + CPr SPC combination with \SI{91.42}{\percent}. The large performance gap between IoU$_{0.5}$ and IoU$_{0.7}$, occurs when the detections are imprecise and the IoU falls between 0.5 and 0.7.
Using CVSA in combination with the CPr methods achieved the best results for AP@IoU$_{0.7}$. For CVSA + CPr SPC, the highest AP with \SI{79.02}{\percent} was achieved. For this method, keypoints are sampled at the locations of collective detections using both SPC for VSA I and SCB for VSA II, resulting in the best performance.

The assumption that more methods in combination always lead to a better perception is refuted by the investigation of a combination of PD, CVSA and CPr strategies. The combination of these three methods performed worst with only \SI{72.36}{\percent} for CPr RoI-grid and \SI{67.66}{\percent} for CPr SPC. This worse performance can arise due to the increased model complexity, which might need additional parameter tuning, or due to correlations that negatively affect the detection performance.

\begin{table}
\centering
\caption{3D average precision results of our proposed fusion together with late fusion}
\label{tab:results-lf}
\renewcommand{\arraystretch}{1.5}
\resizebox{\columnwidth}{!}{%
\begin{tabular}{@{}lll@{}}
\toprule
Method                                    & AP@IoU$_{0.7}$ & AP@IoU$_{0.5}$     \\ \midrule
Late Fusion (HM + WBF)                    & 74.08          & 91.94              \\
RBF + CPr RoI-grid + Late Fusion          & 85.40          & 98.59              \\
RBF + CPr SPC + Late Fusion               & \textbf{85.53} & \textbf{98.63}     \\
PD + CPr RoI-grid + Late Fusion           & 72.86          & 89.80              \\
PD + CPr SPC + Late Fusion                & 73.23          & 89.76              \\
CVSA + CPr RoI-grid + Late Fusion         & 80.97          & 89.54              \\
CVSA + CPr SPC + Late Fusion              & 82.77          & 89.51              \\
PD + CVSA + CPr RoI-grid + Late Fusion    & 76.48          & 85.59              \\
PD + CVSA + CPr SPC + Late Fusion         & 81.80          & 90.45              \\
\bottomrule
\end{tabular}%
}
\end{table}
To rate the results of our proposed approach we conducted a regular late fusion using HM with WBF. Moreover, we combined our CPV-RCNN approach with a subsequent late fusion. The results are presented in Table~\ref{tab:results-lf}. The late fusion reached \SI{74.08}{\percent} for AP@IoU$_{0.7}$ which is slightly better compared to the combined results of the methods including PD or RBF, however, the CVSA + CPr methods could outperform the late fusion by 4.41 p.p. and 4.94 p.p. for AP@IoU$_{0.7}$.
Including the late fusion to the combination of PD + CPr lead to an AP decrease of about 1 p.p. compared to the method without late fusion and the regular late fusion. 
An increase in AP of about 3.5 p.p. compared to without late fusion was achieved by combining the late fusion with CVSA + CPr. Therefore, first the CVSA and CPr methods are applied to enhance the local detection. After that, the improved local detections are fused with the cooperative information using HM and WBF. This combined fusion method achieved an AP of up to \SI{82.77}{\percent} for CPr SPC. This approach outperforms the regular late fusion by 6.89 and 8.69 p.p. 
The best result in combination with a late fusion was achieved using the RBF + CPr methods. For both CPr methods an AP of about \SI{85.50}{\percent} for AP@IoU$_{0.7}$ was achieved which is an increase of more than 11 p.p. compared to the common late fusion. To highlight is also the achieved AP for an IoU threshold of 0.5, which is \SI{98.63}{\percent}.

Different to without late fusion the combination of PD + CVSA + CPr was able to gain an increase in AP. Especially, the version with CPr SPC achieved an AP of \SI{81.80}{\percent}, which is an increase of about 7 p.p. compared to the common late fusion and about 14 p.p. compared to without late fusion. However, this increase in AP can be traced back to the late fusion using highly accurate information.

In total a combination of CVSA and CPr achieved the best result for CPV-RCNN with an AP of \SI{79.02}{\percent}, which corresponds to an improvement of 44.76 p.p. towards the baseline local detection and 4.94 p.p. compared to the late fusion. Using the collective detections to improve the local detection and then perform a late fusion of the improved local detection and the detections from the cooperative vehicles achieved the overall best performance with an AP of \SI{85.53}{\percent} for the combination of RBF + CPr SPC.

\addtolength{\textheight}{-1cm}
\section{Conclusion and Future Work}
\label{conclusion_and_future_work}

In this work we demonstrated a novel fusion approach to incorporate collectively detected objects into the local object detection using the PV-RCNN++ LiDAR object detector in order to enhance the local perception capabilities. 
We proposed four methods which are applied to different parts of the PV-RCNN++ architecture to fuse the cooperative information with local point clouds. Each method was capable to increase the average precision compared to standard PV-RCNN++. Moreover, by applying a combination of collective proposals and a collective voxel set abstraction module we were able to outperform a common late fusion and achieved a 3D average precision of \SI{79.02}{\percent}, which is an increase of 4.94 p.p. compared to the late fusion. 
The approach allows to combine the advantages of early and late fusion.
The communicated data for the approach relies on the late fusion approach through which an efficient communication with reduced data size is enabled. However, the integration of the cooperative information into the raw local data in an early fusion manner leads to a higher information value.

For further research the evaluation will be extended to use a realistic communication model for collective perception. Additionally, a more diverse dataset containing varying numbers of cooperative vehicles and different scenarios, such as intersections, will be used for training and evaluation. Especially the robustness of the proposed approach against false detections of cooperative vehicles will be further investigated.





\section*{Acknowledgment}
This work has been partially funded by the German Research Foundation (DFG) in the priority program 1835 under grant BR2321/5-2.



\bibliographystyle{IEEEtran}
\bibliography{literature}

\end{document}